\documentclass[preprint,12pt]{elsarticle}

\usepackage{amssymb}
\usepackage{algorithm}
\usepackage{algorithmic}

\usepackage{amsmath}
\usepackage[T1]{fontenc}
\usepackage[utf8]{inputenc}
\usepackage[croatian]{babel}
\usepackage{url}
\usepackage{booktabs}
\usepackage{xcolor}
\usepackage{graphicx}
\usepackage{array}
\usepackage{adjustbox}

\journal{Neurocomputing}

\begin{document}

\begin{frontmatter}

\title{Label-independent hyperparameter-free self-supervised single-view deep subspace clustering}

\author{Lovro Sindi\v{c}i\'c, Ivica Kopriva}
\address{Division of Computing and Data Science, Ru\dj er Bo\v{s}kovi\'c Institute\\
Bijeni\v{c}ka cesta 54, 10000 Zagreb, Croatia}


\begin{abstract}
Deep subspace clustering (DSC) algorithms face several challenges that hinder their widespread adoption across various application domains. First, clustering quality is typically assessed using only the encoder's output layer, disregarding valuable information present in intermediate layers. Second, most DSC approaches treat representation learning and subspace clustering as independent tasks, limiting their effectiveness. Third, they assume the availability of a held-out dataset for hyperparameter tuning, which is often impractical in real-world scenarios. Fourth, learning termination is commonly based on clustering error monitoring, requiring external labels. Finally, their performance often depends on post-processing techniques that rely on labeled data. To address these limitations, we introduce a novel single-view DSC approach that (i) minimizes a layer-wise self-expression loss using a joint representation matrix; (ii) optimizes a subspace-structured norm to enhance clustering quality; (iii) employs a multi-stage sequential learning framework, consisting of pre-training and fine-tuning, enabling the use of multiple regularization terms without hyperparameter tuning; (iv) incorporates a relative error-based self-stopping mechanism to terminate training without labels; and (v) retains a fixed number of leading coefficients in the learned representation matrix based on prior knowledge. We evaluate the proposed method on six datasets representing faces, digits, and objects. The results show that our method outperforms most linear SC algorithms with carefully tuned hyperparameters while maintaining competitive performance with the best-performing linear approaches.
\end{abstract}

\begin{keyword}
deep subspace clustering\sep label independent learning\sep hyperparameters free learning\sep self-supervised learning. 


\end{keyword}

\end{frontmatter}


\section{Introduction}

Clustering or partitioning data into disjoint homogeneous groups is one of the fundamental problems in data analysis \cite{1}. It aims to infer structure from data based on similarity between data points. This is relevant to many applied problems, such as image segmentation \cite{2}, data mining \cite{3}, and voice recognition \cite{4}. Since sample spaces often have arbitrary shapes, distance-based algorithms fail to cluster data in the original ambient domain. Moreover, the high dimensionality of the ambient domain further deteriorates performance, a phenomenon known as the \textit{curse of dimensionality}. However, it turns out that data in many applications often have fewer degrees of freedom than the ambient dimension \cite{5}. Due to redundancy, many of the dimensions are irrelevant, i.e., the intrinsic dimension is much smaller. Consequently, the identification of the low-dimensional structure of signals or data in a high-dimensional ambient space is one of the fundamental problems in engineering and mathematics \cite{6}. For instance, principal component analysis (PCA) \cite{7}, independent component analysis (ICA) \cite{8}, and nonnegative matrix factorization (NMF) \cite{9} aim to find a low-dimensional subspace from which all high-dimensional data are assumed to be generated.

However, this model conflicts with practice, where data points are usually drawn from multiple low-dimensional subspaces with unknown membership. Therefore, modeling data using a single low-dimensional subspace is often too restrictive \cite{10}. Subspace clustering addresses this problem by modeling data as a union of multiple subspaces \cite{11}. A significant number of papers is devoted to clustering data points according to the linear low-dimensional subspaces they are generated from \cite{13,14,15}. Nevertheless, in the real world, data do not necessarily come from linear subspaces.

As noted in \cite{16}, in the case of face image clustering, reflectance is more likely non-Lambertian, and the pose of the subject often varies. Thus, it is more likely that the faces of one subject lie on a nonlinear manifold rather than on a linear subspace. One way to address such a problem is the formulation of subspace clustering algorithms in a reproducing kernel Hilbert space (RKHS), employing the kernel trick \cite{17}. However, the performance of kernel-based methods is highly dependent on the choice of kernel functions \cite{18}. After many years of research, it is still unclear how to choose the kernel function so that the kernel-induced RKHS fits empirical data \cite{19}.

The Grassmann manifold-based representation has also emerged as a legitimate candidate to address the clustering of data from nonlinear manifolds \cite{20}. As an alternative to kernel-based methods for nonlinear subspace clustering, deep subspace clustering (DSC) networks have emerged \cite{16,18,21,24}. The motivation behind using neural networks for this task is their ability to learn powerful representations, which can be combined with linear subspace clustering algorithms. By learning an appropriate linear embedding, these networks overcome the fundamental limitations of kernel-based approaches. Theoretically, the learned embedding—the representation at the encoder’s output—should align with the union-of-linear-subspaces model. Consequently, linear self-expressive subspace clustering algorithms with subspace-preserving guarantees \cite{12,15} should effectively cluster data in this embedded space according to their subspaces. 

Once training is complete, the representation matrix, often implemented as a fully connected self-expressive layer \cite{16}, is used to construct the data affinity matrix, which is then processed by spectral clustering to assign labels \cite{25}. However, several limitations prevent DSC from being widely adopted, motivating the research in this paper. 

Single-view DSC methods typically utilize only the last embedding layer of the encoder to construct the affinity matrix \cite{26}. As evidenced by the findings in \cite{27}, the integration of information across multiple layers through a multilayer graph constructed by aggregating the Laplacian matrices corresponding to each layer has been shown to substantially enhance clustering performance. Therefore, in the proposed method, we leverage representations from all encoder layers to construct a joint representation matrix that serves as a common self-expressive model across layers, including the raw input data. 

Many DSC algorithms treat representation learning and subspace clustering as separate problems first constructing an affinity matrix and then applying spectral clustering \cite{12,16,21}. To address this, recent self-supervised approaches \cite{24,26,28,29,30,31,32} integrate clustering quality loss into the training process. The proposed DSC network adopts this principle by incorporating the subspace-structured norm \cite{33}, also used in \cite{28,30}, to provide direct feedback to the self-expression module. 

Despite being self-supervised, many DSC algorithms still rely on labeled data for hyperparameter selection \cite{34,35}. Even methods labeled as "self-supervised" often contain multiple hyperparameters, e.g. five in \cite{29,30} whose optimal values depend on the dataset, making them difficult to tune. This is problematic because clustering is inherently an unsupervised task, yet many DSC algorithms assume access to a held-out labeled dataset for hyperparameter tuning \cite{36}. 

To eliminate this dependency, our method performs learning in two sequential stages:
\begin{enumerate}

    \item A pre-training stage, using either reconstruction loss or distance-preserving loss.
    \item Fine-tuning via layer-wise self-expression loss and clustering quality loss.
\end{enumerate}

This staged design allows multiple regularization terms to be incorporated without any need for hyperparameter tuning. 
Many DSC algorithms determine when to terminate training by monitoring clustering accuracy or error, which, once again, depends on external labels. Our proposed DSC method introduces a self-stopping rule based on relative error. By enforcing consistency among layer-wise affinity matrices, our approach ensures that clustering labels across layers progressively align. Consequently, training is automatically halted once the relative error stabilizes and no longer decreases. Another issue with self-expressive DSC models, first analyzed in \cite{39}, is that their formulation can often be ill-posed, meaning that data in the embedded space may not fully adhere to the union-of-linear-subspaces model. This suggests that some of the reported clustering performance improvements may stem from ad hoc post-processing techniques rather than the DSC model itself \cite{39}. In our proposed autonomous DSC method, we optionally post-process the learned representation matrix by retaining only the $d$ leading coefficients, leveraging a priori knowledge of subspace dimension $d$ \cite{40} (see Section 2.1).

To address these challenges, we propose Label-Independent Hyperparameter-Free Self-Supervised Single-View Deep Subspace Clustering (LIHFSS-SVDSC). The code implementation is available at: https://github.com/lovro-sinda/LIHFSS-SVDSC\\

Experiments conducted on six widely used datasets demonstrate that the LIHFSS-SVDSC algorithm consistently outperforms eight linear single-view subspace clustering (SC) algorithms with carefully tuned hyperparameters, sometimes by a significant margin. Furthermore, its clustering performance is comparable to that of single-view deep subspace clustering (DSC) algorithms that rely on hyperparameter tuning.

Our key contributions are summarized as follows:

\begin{enumerate}
    \item We introduce a self-supervised single-view deep subspace clustering (DSC) method that is free of hyperparameters and external labels. The approach employs a sequential learning framework, beginning with pre-training using either reconstruction loss or distance-preserving loss, followed by fine-tuning with layer-wise self-expression loss (utilizing a shared representation matrix) and clustering quality loss.
    
    \item To further improve clustering accuracy, we propose an optional post-processing step that refines the learned representation matrix. This step is grounded in rigorous mathematical properties derived from the low-dimensional subspace model and incorporates a priori knowledge of subspace dimensions.
     \item We introduce a self-stopping criterion based on epoch-wise relative error, which allows the algorithm to terminate training automatically without requiring external labels or predefined stopping rules.
    
    \item Comprehensive experiments on six widely used datasets validate the effectiveness of our approach. The LIHFSS-SVDSC algorithm consistently outperforms eight linear single-view SC algorithms with hyperparameter tuning and demonstrates competitive performance against single-view deep subspace clustering methods that require fine-tuned hyperparameters. Ablation studies further confirm the importance of the proposed loss functions and architectural choices.
\end{enumerate}
Proposed LIHFSS-SVDSC method is illustrated in Figure 1.
\begin{figure}[H]
  \centering
  \label{slika}
  \includegraphics[scale=0.90]{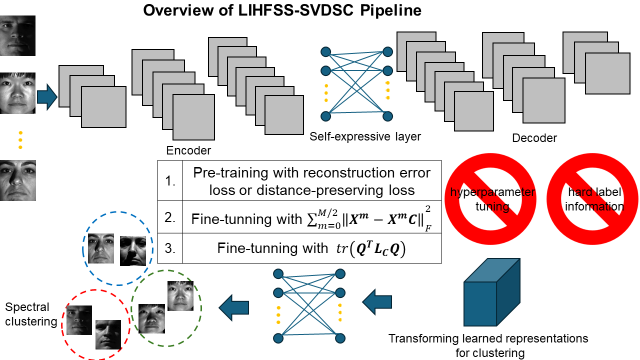}
  \caption{The learning process is organized sequentially in two stages: pretraining stage based on either reconstruction error loss or distance-preserving loss, and fine-tuning stage which combines layerwise self-expression loss and clustering quality loss.}
\end{figure}
\section{Background and Related Work}

Before we formally introduce the subspace clustering problem, let us emphasize that notation used in this paper is introduced clearly when first needed.
Table 1 summarizes some
notation used in the paper.
\begin{table}[H]
\label{tab12}
\caption{Notations and Abbreviations}
\centering
\begin{tabular}{|c|l|}
\hline
\textbf{Notation} & \textbf{Definition} \\ \hline
$N$ & Number of data points \\ \hline
$D$ & Dimension of data points \\ \hline
$K$ & Number of subspaces \\ \hline
$M$ & Number of layers in the autoencoder \\ \hline
$\mathbf{X} \in \mathbb{R}^{D \times N}$ & Data matrix \\ \hline
$\mathbf{C} \in \mathbb{R}^{N \times N}$ & Representation matrix \\ \hline
$\mathbf{W} \in \mathbb{R}^{N \times N}$ & Affinity matrix \\ \hline
$\mathbf{Q} \in \mathbb{N}_0 ^{N\times C}$ & Cluster indicator matrix obtained by spectral clustering \\ \hline

\end{tabular}
\label{table:notations}
\end{table}

\subsection{Subspace Clustering with Self-Expressive Model}

Let us assume that $\mathbf{X} \in \mathbb{R}^{D \times N} := \{\mathbf{x}_1, \mathbf{x}_2, ..., \mathbf{x}_N \}$ represents a collection of $N$ data points (patterns, instances) in a $D$-dimensional ambient space. In general, subspace clustering models assume data points are drawn from $C \geq 1$ affine subspaces:

\begin{equation}
    B_c = \{ \mathbf{x}_n \in \mathbb{R}^{D}: \mathbf{x}_n = \mathbf{A}_c \mathbf{b}_n^c + \boldsymbol{\mu}_c \}, \quad c \in \{1, ..., C\}, \quad n \in \{1, ..., N\}
\end{equation}

where the dimensions $\{d_c < D\}_{c=1}^{C}$ \cite{11,12}. $\{ \mathbf{A}_c \in \mathbb{R}^{D \times d_c} \}_{c=1}^{C}$ represent low-dimensional subspace bases and $\mathbf{b}_n^c\in\mathbb{R}^{d_c}$  represents the low-dimensional representation of data point $\mathbf{x}_n$ in subspace $c \in \{1, \dots, C\}$. When $\boldsymbol{\mu}_c = 0$, subspaces are linear. It was shown in \cite{41} that when the dimension of the ambient space is high relative to the sum of the dimensions of affine subspaces, i.e., $ \sum_{c=1}^{C} d_c < D$, the affine constraint has negligible influence on clustering performance. Thus, in the rest of the paper, we shall assume the model with linear subspaces in (1). As is common in practice, we assume that the number of subspaces $C$ is equivalent to the number of clusters, which is known \textit{a priori}. In many subspace clustering algorithms, only data clustering without explicit identification of subspace bases and subspace dimensions is required \cite{12,15}. Various subspace clustering algorithms first focus on learning a weighted directed graph representing interactions between data points. Afterward, spectral clustering \cite{25} is applied to the learned graph to cluster data points according to the subspaces they are drawn from.

Let us represent the data matrix as:
$    \mathbf{X} = [\mathbf{X}_1, \mathbf{X}_2, ..., \mathbf{X}_C]\mathbf{T},
$
where $\mathbf{T}$ is an arbitrary permutation matrix. Without loss of generality, we assume $\mathbf{T} = \mathbf{I}$. Based on (1), and assuming the presence of error, we have the following representation of our dataset using the linear subspace model:
\begin{equation}
    \mathbf{X} = \mathbf{A} \mathbf{B} + \mathbf{E}
\end{equation}

where $\mathbf{A} = [\mathbf{A}_1, ..., \mathbf{A}_C]$, $\mathbf{B}$ is a block diagonal matrix with blocks on the main diagonal $\{\mathbf{B}_c \in \mathbb{R}^{d_c \times N_c} \}_{c=1}^{C}$, where $\bigcup_{c=1}^{C} N_c = N$, and $\mathbf{E}$ represents the error term. 

The data affinity matrix, which reflects the multiple subspace structure of data points, is the main challenge in subspace clustering algorithms. It is often built by exploiting the self-expression property \cite{12,13,15}, which uses the fact that each data point can be represented as a linear combination of other data points from the same subspace, as in model (1). We obtain the self-expressive model from (2) by setting $\mathbf{A} = \mathbf{X}$ and formally $\mathbf{B} = \mathbf{C}$:

\begin{equation}
    \mathbf{X} = \mathbf{X} \mathbf{C} + \mathbf{E}.
\end{equation}

In (3), $\{ c_{ij} \}_{i,j=1}^{N}$ expresses similarity between corresponding data points $(\mathbf{x}_i, \mathbf{x}_j)$. If $(\mathbf{x}_i, \mathbf{x}_j)$ belong to the same cluster, $c_{ij}$ should be large; otherwise, it should be small. The matrix $\mathbf{C}$ in (3) can also be interpreted as a new representation of data $\mathbf{X}$ in frame $\mathbf{X}$. The self-expressive model-based subspace clustering algorithms use various optimization approaches to learn a good representation matrix $\mathbf{C}$ from data. Once it is learned, we obtain the data affinity matrix as:

\begin{equation}
    \mathbf{W} = \frac{|\mathbf{C}| + |\mathbf{C}^T|}{2}.
\end{equation}

Let us define the diagonal degree matrix as $\mathbf{D} = \operatorname{diag}(\mathbf{d_i})$ and $\{d_{ii} =\sum_{m=1}^{N} w_{im}\}_{i=1}^{N}$,
where $\mathbf{L}$ is the graph Laplacian matrix, defined as: $\mathbf{L} = \mathbf{D} - \mathbf{W}$.
The normalized form of $\mathbf{L}$ is given in \cite{42}:
\begin{equation}
    \mathbf{\tilde{L}} = \mathbf{D}^{-1/2} (\mathbf{D} - \mathbf{W}) \mathbf{D}^{-1/2} = \mathbf{I} - \mathbf{D}^{-1/2} \mathbf{W} \mathbf{D}^{-1/2}.
\end{equation}

The matrix $\mathbf{\tilde{L}}$ is symmetric and positive semi-definite with eigenvalues that lie in $[0,2]$. The binary cluster indicator matrix $\mathbf{F} \in \mathbb{N}_0^{N \times C}$ is obtained by applying k-means clustering on $C$ eigenvectors of $\mathbf{\tilde{L}}$ corresponding to the $C$ smallest eigenvalues. The shifted Laplacian matrix is defined as \cite{42}:

\begin{equation}
    \mathbf{L}_s = 2\mathbf{I} - \mathbf{\tilde{L}} = \mathbf{I} + \mathbf{D}^{-1/2} \mathbf{W} \mathbf{D}^{-1/2}.
\end{equation}

According to property 3 in \cite{42}, if $(\lambda, \mathbf{v})$ is an eigenvalue-eigenvector pair of $\mathbf{\tilde{L}}$, then $(2 - \lambda, \mathbf{v})$ is an eigenvalue-eigenvector pair of $\mathbf{L}_s$. Thus, we can obtain the cluster indicator matrix $\mathbf{F}$ by applying k-means clustering on $C$ eigenvectors of $\mathbf{L}_s$ corresponding to the $C$ largest eigenvalues. This makes the clustering approach based on the shifted Laplacian more robust to the presence of noise. All results reported in Section 4 are based on the shifted Laplacian.

The representation matrix $\mathbf{C}$ is subspace-preserving if for any data point $\mathbf{x}_i \in B_j$, $c_{ij} = 0$ for all $j$ when $\mathbf{x}_i \notin B_j$. To that end, appropriate regularization has to be imposed on $\mathbf{C}$ when it is estimated from data. Generally, the linear subspace clustering problem based on the self-expressive property is formulated as follows:

\begin{equation}
    \min_{\mathbf{C} \in \mathbb{R}^{N \times N}} f(\mathbf{C}) + \lambda g(\mathbf{E}) \quad \text{s.t.} \quad \mathbf{X} = \mathbf{X} \mathbf{C} + \mathbf{E}, \quad \text{diag}(\mathbf{C}) = \mathbf{0}
\end{equation}

where $f$ represents the regularization function, and $g$ is the function that models noise (e.g., $\ell_1$ norm for sparse gross noise \cite{13}, the Frobenius norm for Gaussian noise \cite{43}, $\ell_{2,1}$ norm for specific corruptions (outliers) \cite{12}). The parameter $\lambda > 0$ is a tradeoff parameter. The constraint $\text{diag}(\mathbf{C}) = 0$ prevents data points from being represented by themselves. Fulfillment of the subspace-preserving condition strongly depends on regularization imposed on $\mathbf{C}$.  Some choices are: $\|\mathbf{C}\|_1$ in \cite{13} or $\|\mathbf{C}\|_*$ in \cite{12}. For more details regarding subspace-preserving guarantees, see Table 1 in \cite{19}. From the optimization point of view, the problem (7) is solved efficiently using the alternating direction method of multipliers (ADMM) \cite{44}.

Regarding the selection of error regularization function $g(\mathbf{E})$, the problem is that the structure of the error should be known in the form of prior knowledge. In real-world datasets, that assumption is not always fulfilled. To address this, an approach is proposed in \cite{40} that eliminates errors in projection terms, especially when the dimension of the original ambient space $D$ is particularly high. As a result, $\ell_{1}$, $\ell_{2,}$, $\ell_{\infty}$ and nuclear-norm-based linear projection spaces share the same properties of intra-subspace projection dominance (IPD). In other words, coefficients over intra-subspace data points are larger than those over inter-subspace data points. Thus, after optimization (7) is finished, the obtained representation matrix $\mathbf{C}$ is post-processed by keeping the $d_c$ largest magnitude coefficients. Here, $d_c$ stands for the dimension of the subspace to which a data point belongs. As already explained, we shall assume that all the subspaces have the same unknown subspace dimension, i.e., $\{d_c = d\}_{c=1}^{C}$. 

In many scenarios, the dimension of the subspaces, $d$, is known. As an example, face images of each subject in the Yale B dataset lie approximately in a $d=9$ subspace \cite{45}. Handwritten digits, such as those in the MNIST and USPS datasets, lie approximately in a $d=12$ subspace \cite{46}. Regarding object-based datasets, such as COIL20 and COIL100, the recommended subspace dimension is $d=9$ \cite{47}. 

Thus, in the proposed method, post-processing of the learned representation matrix can be performed as:

\begin{equation}
\mathbf{C} \leftarrow \left\lceil \mathbf{C} \right\rceil_{d}
\end{equation}

without introducing a new hyperparameter.

\subsection{Deep Subspace Clustering with Self-Expressive Model}

As opposed to the linear model (1), nonlinear subspace clustering presumes that data points lie on $C$ manifolds \cite{19}:

\begin{equation}
    \mathcal{M}_c = \left\{ \mathbf{x}_n \in \mathbb{R}^{D \times 1} : \mathbf{x}_n = f_c \left( \mathbf{A}_c\mathbf{b}_n^c \right) \right\}_{n=1}^{N}, \quad c \in \{1, ..., C\}
\end{equation}

where $f_c$ is the unknown nonlinear manifold mapping function from $\mathbb{R}^{d_c}$ to $\mathbb{R}^{D}$. The problem of nonlinear subspace clustering, also known as nonlinear manifold clustering, is defined as segmenting data points according to their corresponding manifolds and obtaining the low-dimensional embedding $\mathbf{b}_n^c$ of each data point $\mathbf{x}_n$ within each manifold \cite{49}. The main motivation for DSC, in terms of neural networks, was to merge their powerful representation learning capability with the linear subspace clustering algorithms to deal with data generated from nonlinear manifolds.
Before proceeding further, let us introduce the following notation for outputs of encoder layers:
$    \{\mathbf{X}^m \in \mathbb{R}^{D_m \times N} \}_{m=0}^{M/2}
$ assuming the encoder has $M/2$ layers and also assuming $\mathbf{X}^0 = \mathbf{X}$. The group of DSC algorithms based on self-expressive latent representation learning \cite{16,18,22} are based on the following optimization problem:

\begin{equation}
    \min_{\mathbf{C}, \Theta} \|\mathbf{X}^{M/2}_\Theta - \mathbf{X}^{M/2}_\Theta \mathbf{C} \|_F^2 + \lambda_1 f(\mathbf{C}) + \lambda_2 h(\mathbf{X},\mathbf{X}^{M/2}_\Theta, \mathbf{C})
\end{equation}
\[
    \text{such that } \quad \mathbf{X}^{M/2}_\Theta = \Phi_{e} (\mathbf{X}, \Theta) \quad \text{and} \quad \text{diag}(\mathbf{C}) = \mathbf{0}
\]

where $\Phi_{e}$ denotes network embedding, the output of the encoder with parameters $\mathbf{\Theta}$ and input $\mathbf{X}$. By minimizing the self-expressive representation term, the latent representation $\mathbf{X}^{M/2}_\Theta$ is encouraged to obey the union-of-linear-subspaces structure, and $\Theta$ represents the network’s parameters. The function $f(\mathbf{C})$ imposes regularization on the representation matrix, and $h(\mathbf{X}, \mathbf{X}^{M/2}_\Theta, \mathbf{C})$ plays a critical role in removing trivial solutions and specifying properties of nonlinear mapping and embedding space. For example, 
$h(\mathbf{X}^{M/2}_\Theta) = \frac{1}{2} \sum_{i,j} \|\mathbf{X}^{M/2}_\Theta(:,i) - \mathbf{X}^{M/2}_\Theta(:,j)\|_1^2,$
in \cite{18} prevents arbitrary scaling factors in the embedding space. 

Two constraints were imposed on the representation matrix in \cite{16}: $f(\mathbf{C}) = \|\mathbf{C}\|_1$ and $f(\mathbf{C}) = \|\mathbf{C}\|_{2}$, leading to DSC-L1 and DSC-L2 networks. The network in \cite{22} imposes $h(\mathbf{X}, \mathbf{C}) = \frac{1}{2} \operatorname{tr} (\mathbf{C} \mathbf{L} \mathbf{C}^T)$
where $\mathbf{L}$ is the Laplacian matrix formulated on the basis of the adjacency matrix over the original dataset $\mathbf{X}$. Thus, imposed regularization preserves the local structure of data. It is motivated by the local invariance assumption \cite{37,38}. 

The work in \cite{16} was the first to formally replace the self-expressive term in (10) by a fully connected layer, i.e., coefficients of $\mathbf{\Theta_s}$ formally substituted by the network parameters $\Theta$. In other words, we now have $\mathbf{X}^{M/2}_{\Theta_e} = \mathbf{X}^{M/2}_{\Theta_e} \mathbf{\Theta_s}$, and $\Theta_e$ stands for parameters of the encoder. 

The network is now parameterized in terms of $\mathbf{\Theta} := \{\mathbf{\Theta_e}, \mathbf{\Theta_s}, \mathbf{\Theta_d\}}$, where $\mathbf{\Theta_d}$ stands for decoder parameters. The learning problem of DSC network \cite{16} is formulated as:

\begin{equation}
    \min_{\Theta} \left\| \mathbf{X} - \tilde{\mathbf{X}}_{\Theta} \right\|_F^2 
+ \lambda_1 \left\| \Theta_s \right\|_p 
+ \frac{\lambda_2}{2} \left\| \mathbf{X}_{\Theta_e}^{M/2} - \mathbf{X}_{\Theta_e}^{M/2} \Theta_s \right\|_F^2 
 \text{, such that } \operatorname{diag}(\Theta_s) = \mathbf{0}
\end{equation}

where $p \in \{1,2\}$. Thus, DSC-Net-L1 and DSC-Net-L2 were implemented in \cite{16}. The network (11) is trained in two stages. First, pre-training using reconstruction error only, and second fine-tuning which includes the self-expressive layer. Once trained, the representation matrix $\mathbf{C} = \Theta_s$ is used to build the Laplacian matrix to be used by spectral clustering. 
Recent studies have explored variational deep embedding for unsupervised clustering. Notably, a decoder-free approach was proposed in \cite{decoderFreeClustering2024}, demonstrating improved latent space organization. Our work extends this by integrating self-supervised distance-preserving loss without requiring pre-defined hyperparameters.

\subsection{Self-Supervised Deep Subspace Clustering}

As can be seen in (10) and (11), DSC algorithms contain hyperparameters in terms of regularization constants. To achieve good clustering performance, these hyperparameters need to be tuned carefully for each dataset, which requires a held-out dataset. For many scenarios, such an approach is not feasible. Herein, we elaborate on several DSC algorithms that operate in a label-independent manner.

In \cite{21}, a deep embedded clustering method is proposed. It uses a deep neural network to \textit{simultaneously} learn feature representations and cluster assignments. After the autoencoder is pre-trained, the network minimizes Kullback-Leibler (KL) divergence between soft labels and the auxiliary distribution. Soft labels are obtained by measuring similarity between the encoder output and cluster centroids. Auxiliary distributions are based on estimated soft labels. Gradients of KL divergence, which represent the loss function w.r.t. encoder outputs and centroids, are used to guide the network.

In \cite{35}, the deep embedded regularized clustering (DEPICT) algorithm is used for simultaneous representation learning and cluster assignment. The DEPICT algorithm is also self-supervised, making it suitable for real-world data clustering tasks where no label data is available for hyperparameter tuning. It simultaneously trains all encoder and decoder layers together with the softmax layer that performs cluster assignment. The loss function is composed of KL divergence between model predictions and target variables, and a term that represents errors between corrupted encoder layers and their current reconstruction. Thereby, the loss function is free of hyperparameters, making the DEPICT algorithm suitable for real-world clustering problems, where no held-out data are required for tuning the hyperparameters.

In \cite{24}, a pseudo-supervised deep subspace clustering (PSSC) method is developed. The method is built on the idea of self-training (self-supervision, self-guidance) through multiple training iterations. At each iteration, the model uses predictions from the last iteration to relabel unlabeled samples. This is achieved by attaching the classification module to the encoder output. It uses the latent representation $\mathbf{Z}_e$ and the similarity graph to construct pseudo-labels used to guide feature learning and graph learning \cite{50}.
The overall loss function of the PSSC method contains three hyperparameters. They are fine-tuned by using the grid search, i.e., labeled data are necessary to select the optimal values of the hyperparameters. Proposed LIHFSS-SVDSC algorithm is in contrast to some extent comparable with \textit{DeepClue}, a unified deep clustering framework that jointly explores features learned in multiple network layers for enhancing clustering performance \cite{50}. That is achieved by utilizing a weight-sharing CNN as the backbone to learn representations of the sample pairs constructed by different data augmentations. Furthermore, two two-layer perceptrons were used for instance-level projectors and cluster-level projectors. The first one projects embedded instances into a new embedded space where instant-level contrastive loss is calculated. The second one projects embedded instances via a \textit{softmax} layer into cluster space. There, cluster-level contrastive loss is calculated.

A number of layers are extracted from the backbone, instance projector, and cluster projector. From them, the ensemble of base clustering is generated by the ultra-scalable spectral clustering (U-SPEC) algorithm that has linear time and space complexity \cite{51}. Our approach differs from \textit{DeepClue} in a way that we use all the encoder layers, including the data from the original input space, to construct a target distribution. That enforces layer-wise cluster assignments to be globally consistent. Although the proposed LIHFSS-SVDSC method is focused on single-view data, it is also to a certain extent comparable to self-guided deep multi-view subspace clustering (SDMSC) model \cite{26}. SDMSC performs joint deep feature embedding and subspace analysis of multi-view data and estimates a consensus-data affinity relationship agreed upon by features from not only all views but also at intermediate embedding spaces. Thus, it coincides with our goal herein. However, SDMSC ensures effective deep feature embedding without label supervision by using the data affinity relationship obtained with raw features as the supervision signals to self-guide the embedding process. In other words, the representation matrix estimated from the input raw data by the linear sparse subspace clustering algorithm \cite{13}, is used as a target for layer-wise data representation matrices. Even though the data affinity relationship based on raw features is imprecise, it preserves the new intrinsic structure of data samples. In the proposed LIHFSS-SVDSC algorithm, we focus on single-view data and do initial learning in the pre-training stage, minimizing either reconstruction error loss or distance-preserving loss. Specifically, distance-preserving loss is capable of preserving the intrinsic structure of data. Afterward, the consensus representation matrix is learned in the fine-tuning stage, minimizing layer-wise self-expression loss and clustering quality loss.

The main motivation for the development of self-supervised DSC is enabling deep neural networks to utilize unlabeled data by removing excessive annotated labels. In that regard, \cite{29} and \cite{30} stand for two instructive examples of self-supervised clustering networks. The self-supervised convolutional subspace clustering network (S$^2$ConvSCN) \cite{29} combines convolutional \textit{autoencoder} with a fully connected (FC) layer and a self-expressive layer $\Theta$, with a spectral clustering module. The FC layer is trained on features extracted by the encoder and generates the so-called soft-labels. Spectral clustering generates the pseudo-labels, in terms of one-hot coding vectors, that are based on the learned representation matrix $\Theta$. The total loss function of the S$^2$ConvSCN is composed of five losses:

\begin{equation}
    \mathcal{L} = \mathcal{L}_{\text{RE}} + \lambda_c \mathcal{L}_{\text{C}} + \lambda_s \mathcal{L}_{\text{LSE}} + \lambda_q \mathcal{L}_Q + \lambda_{\text{LCE}} \mathcal{L}_{\text{LCE}}
\end{equation}

where:

\begin{equation}
    \mathcal{L}_{RE} = \frac{1}{2N} \left\| \mathbf{X} - \tilde{\mathbf{X}}_{\Theta} \right\|_F^2\end{equation}

stands for the reconstruction error loss, $\mathcal{L}_{\text{C}} = \|\mathbf{C}\|_1$ stands for the regularization loss imposed on representation matrix $\mathbf{C}=\Theta_s$, $
    \mathcal{L}_{\text{LSE}} = \frac{1}{2} \|\mathbf{X}^{M/2}_{\Theta_e} - \mathbf{X}^{M/2}_{\Theta_e} \mathbf{C}\|_F^2$ stands for the self-expression error loss, and 
    
\begin{equation}
\mathcal{L}_{\mathbf{Q}} = \sum_{i,j=1}^{N} \left| c_{ij} \right| \frac{\left\| \mathbf{q}_i - \mathbf{q}_j \right\|_2^2}{2}
= \operatorname{tr} \left( \mathbf{Q}^\top \mathbf{L}_C \mathbf{Q} \right) =: \left\| \mathbf{C} \right\|_{\mathbf{Q}}
\end{equation}

where $\mathbf{Q} \in \mathbb{N}_0 ^{N\times C}$, $\mathbf{Q}^T \mathbf{Q} = \mathbf{I}_C$, is the cluster indicator matrix obtained by spectral clustering, and $\mathbf{L}_C$ is the Laplacian built on the learned representation matrix $\mathbf{C}$. In (14), it is used that $w_{ij} = (|c_{ij}| + |c_{ji}|)/2.$ Thus, minimizing (14) enforces $\mathbf{C}$ such that $c_{ij} \neq 0$ only if $\mathbf{x}_i$ and $\mathbf{x}_j$ have the same label. This term is called the self-supervised prior in \cite{31}. In addition to self-expressive layers, $\mathbf{Q}$ also provides feedback to the encoder supervisor. Encoder supervisor is implemented in terms of the FC layer with the encoder output as its input. Based on representations $\left\{ \mathbf{z}_{\Theta_e}^{n} \right\}_{n=1}^{N}
$, the supervisor tries to predict their labels $\left\{ \tilde{\mathbf{y}}_n \in \mathbb{R}^{C \times 1} \right\}_{n=1}^{N}
$ that are termed soft labels. To preserve the meaning of probability, the \textit{softmax} layer is placed at the output of the FC supervisor layer.
Supervision of feature extraction process is implemented through minimization of the loss $\mathcal{L}_{\text{CE}}$ that combines cross-entropy loss and center loss. Self-supervised deep subspace clustering method in \cite{31} is based on graphs but exploits ideas of S$^2$ConvSCN with an extra loss term. The critique in \cite{39} related to deep networks in general is also applicable to self-supervised deep subspace clustering models \cite{30,31}. 

Even though they do not require labeled data for learning, their performance depends on many hyperparameters, five in the case of \cite{30,31}. Their optimal values are typically dataset dependent, making the selection process demanding. It is actually always assumed that a held-out dataset is available to select the hyperparameters in grid-search approach through maximum accuracy or minimum clustering error. S$^2$ConvSCN even uses clustering accuracy as a criterion to stop the learning process. Moreover, it uses clustering accuracy to select a post-processing threshold that sets the small values of learned representation matrix $\mathbf{\Theta_s}$ to zero. This assumption contradicts the unsupervised nature of the task.

\section{The Proposed Deep Subspace Clustering Method}

Herein, we present a label-independent, hyperparameter-free, self-supervised, single-view deep subspace clustering (LIHFSS-SVDSC) algorithm. The algorithm is designed to implement DSC for single-view data in a truly autonomous manner in the spirit of the DEPICT algorithm \cite{39} and SDMSC algorithm \cite{26}. 

Thereby, as opposed to \cite{39} and \cite{26}, learning of LIHFSS-SVDSC is organized in a sequential way, starting with pre-training and proceeding with fine-tuning in multiple stages. Let us now suppose we want to implement learning with multiple loss functions in the spirit of (12) as:

\begin{equation}
    \mathcal{L} = \mathcal{L}_{0} + \lambda_1 \mathcal{L}_{1} + \lambda_2 \mathcal{L}_{2}.
\end{equation}

To avoid problems associated with hyperparameter tuning, we proceed as follows. The $\mathcal{L}_0$ loss is supposed to be minimized during the pre-training stage. Reconstruction error (13) is one choice selected by many DSC algorithms, i.e., $\mathcal{L}_0 = \mathcal{L}_{\text{RE}}$. By minimizing reconstruction error loss, the autoencoder aims to achieve compressive representation \cite{18}, in which case block-diagonality of data affinity matrix is compromised. 

To that end, in addition to reconstruction error loss, we also propose using distance-preserving loss in the pre-training stage. Thus, we formulate the weighting matrix in the original ambient space as:

\begin{equation}
  \left\{ \mathbf{H}_{ij} = \left\| \mathbf{x}_i - \mathbf{x}_j \right\|_{2} \right\}_{i,j=1}^{N}
\end{equation}

and distance-preserving loss as:

\begin{equation}
    \mathcal{L}_0 = \mathcal{L}_{\text{DP}} = \sum_{m=0}^{M/2} \sum_{i,j=1}^{N} \mathbf{H}_{ij} \|\mathbf{x}_i^m - \mathbf{x}_j^m\|_2.
\end{equation}

As elaborated before, the use of distance-preserving loss is motivated by the local invariance assumption \cite{37,38}. It preserves the original structure of data such that the belonging of data points to clusters should be invariant to representations. That is essentially equivalent to one of the goals of the proposed LIHFSS-SVDSC algorithm when striving to learn a consensus-data-affinity relationship agreed by features from all intermediate embedding spaces (layers). Thus, in relation to (15), we want to minimize in the fine-tuning stage the $\mathcal{L}_1$ loss:

\begin{equation}
    \min_{\mathbf{C}} \mathcal{L}_1 = \mathcal{L}_{\text{SE}} = \sum_{m=0}^{M/2} \|\mathbf{X}^{m} - \mathbf{X}^{m} \mathbf{C}\|_F^2.
\end{equation}

As already pointed out, many DSC algorithms treat representation learning and subspace clustering as two independent problems. To address this issue, many self-supervised, also known as self-guided, algorithms are developed \cite{24,26,28,32}. Thus, in the proposed DSC network, we also integrate clustering quality loss:

\begin{equation}
    \min_{\mathbf{Q}} \mathcal{L}_2 = \mathcal{L}_{Q}
\end{equation}

in the second fine-tuning stage. In (19), clustering quality loss $\mathcal{L}_Q$ is given in (14). When learning is finished, the representation matrix is obtained as:

\begin{equation}
    \mathbf{C} = \Theta_s.
\end{equation}

We are aware that the presented sequential multi-stage learning is not equivalent to simultaneously minimizing the overall loss (15). However, the proposed approach eliminates the use of hyperparameters and provides truly autonomous learning. It is, however, important to notice that in (14), $\mathcal{L}_{Q} = \operatorname{tr}(\mathbf{Q}^T \mathbf{L}_C \mathbf{Q}),$ i.e., minimization with respect to $\mathbf{Q}$ preserves local relationships established during the first fine-tuning stage. That explains the results of the ablation study, see section 4.4, in which the second fine-tuning stage still improved clustering performance. Analogously, if we reverse the fine-tuning sequence, minimization of $\mathcal{L}_Q$ first will preserve local structures
relationships established during pre-training stage, especially if distance-preservation loss is
used. Afterwards, minimization of \(\mathcal{L}_{\text{RE}}\) (18) will also preserve local relationships established during first fine-tuning stage. The reason is that learned consensus representation matrix in (18)
is agreed by features, representations, by all intermediate embedding spaces including the input data $\mathbf{X}$. See results in sections 4.3 and 4.4 that support this statement.
To ensure that the proposed LIHFSS-SVDSC algorithm operates in a fully autonomous and label-independent manner, we employ the concept of relative error. Since this error is inherently unsupervised, it serves as a natural stopping criterion. Specifically, we define \(\varepsilon_t\) as the reconstruction error or discrepancy measure at iteration \(t\), which quantifies the difference between successive affinity matrices. The algorithm terminates when \(\varepsilon_t\) ceases to decrease or when its relative change falls below a predefined threshold:

\begin{equation}
    \frac{\varepsilon_t - \varepsilon_{t-1}}{N} \leq \delta.
\end{equation}

In our experiments, we set \(\delta=0.01\) in all experiments reported in section 4. We summarize
proposed LIHFSS-SVDSC algorithm in Algorithm 1.
\pagebreak
\begin{algorithm}
    \caption{Label-independent hyperparameter-free self-supervised single-view deep subspace clustering (LIHFSS-SVDSC)}
    \textbf{Input:} dataset \(\mathbf{X} \in \mathbb{R}^{D \times N}\), number of clusters \(C\).
    \begin{enumerate}
        \item Pre-training: minimizing \(\mathcal{L}_{\text{RE}}\) (13) or \(\mathcal{L}_{\text{DP}}\) (17).
        \item Fine-tuning by (18).
        \item Fine-tuning by (19).
        \item Terminate learning when (21) is satisfied.
        \item Obtain representation matrix \(\mathbf{C}\) from (20).
        \item Optionally post-process representation matrix using (8).
        \item Compute data affinity matrix using (4).
        \item Compute shifted Laplacian matrix using (6).
        \item Apply spectral clustering to shifted Laplacian matrix to assign labels to data.
    \end{enumerate}
    \textbf{Output:} Assigned labels $\mathbf{F}\in \mathbb{N}_0^{N\times C}$
\end{algorithm}

\section{Experiments}
\subsection{Datasets and Clustering Quality Metrics}
We performed comparative performance analysis on six widely known datasets: MNIST \cite{52},
USPS \cite{53}, Extended YaleB \cite{54}, ORL \cite{55}, COIL20, and COIL100 \cite{56}. The main information on
datasets used in the experiments is summarized in Table 2. MNIST and USPS contain digit
images, ORL and Extended YaleB contain face images, and COIL20 and COIL100 contain images of
objects.

\begin{table}[h]
    \centering
    \caption{Main information on datasets used in the experiments.}
    \begin{tabular}{lccc}
        \toprule
        Dataset & Samples (\(N\)) & Features (\(D\)) & Clusters (\(C\))  \\
        \midrule
        MNIST & 10000 & 28$\times$28 & 10  \\
        USPS & 7291 & 16$\times$16 & 10  \\
        Extended YaleB & 2432 & 48$\times$42 & 38 \\
        ORL & 400 & 32$\times$32 & 40 \\
        COIL20 & 1440 & 32$\times$32 & 20 \\
        COIL100 & 7200 & 32$\times$32 & 100 \\
        \bottomrule
    \end{tabular}
    \label{tab:datasets}
\end{table}

Regarding clustering quality metrics, we selected accuracy (ACC), normalized mutual
information (NMI), and \(F_1\)-score. They are widely accepted metrics for measuring quality of
clustering \cite{12,18,21,24,26,32}. All three measures lie between 0 and 1, where 0 means
the worst performance and 1 means the best performance.
All experiments were run in Python 3.12 with Pytorch and Numpy environment on the
PC with 51 GB of RAM and Intel Xeon CPU processors operating with a clock speed of 2 GHz.

\subsection{Algorithms to be compared}
The proposed LIHFSS-SVDSC algorithm is fully autonomous. It is hyperparameter free and, therefore, does not require held out dataset for hyperparameters tuning. It is self-supervised, i.e., employed loss
functions do not use labels also. The self-stopping rule for termination of learning is based on the fully unsupervised relative error concept. Taking everything into account, our intention
is to design a fair comparative performance analysis. Thus, we excluded multi-view SC algorithms because through selection of views and features, a lot of a priori information is
helping them in achieving good performance. We included both the RE loss and DP loss of the LIHFSS-SVDSC algorithm, applying all loss functions simultaneously with optimized hyperparameters, treating them as oracles rather than baselines.
We also included
linear single-view SC algorithms \cite{12,13,15,57,58,59} with carefully tuned
hyperparameters in comparative performance analysis. 

Low-rank representation (LRR) SC \cite{12}
learns the low-rank representation matrix $\mathbf{C}$ in the self-representation data model, where the
low-rank constraint is implemented in terms of the $S_1$ norm. The sparse SC (SSC) algorithm
\cite{13} can be run in two modes: assuming additive white Gaussian noise and assuming outliers.
It learns a representation matrix by imposing an $\ell_1$-norm based sparsity constraint on the
representation matrix. 

Generalization of minimax concave penalty low-rank sparse SC (GMC-
LRSSC) \cite{15} and $S_0$-$\ell_0$ low-rank sparse SC (S0L0-LRSSC) \cite{15} impose simultaneously low-
rank and sparsity constraints on the representation matrix $\mathbf{C}$. Nearest subspace neighbor (NSN)
algorithm \cite{58} first determines the neighborhood set of each data point and then uses a greedy
subspace recovery algorithm to estimate the subspace from the given set of points. The robust
thresholding SC (RTSC) algorithm \cite{57} estimates the adjacency matrix by estimating the set
of $q$ nearest neighbors for each data point according to the metric
\begin{equation}
    s(\mathbf{x}_i, \mathbf{x}_j) = \arccos\left( \frac{\langle \mathbf{x}_i, \mathbf{x}_j \rangle}{\|\mathbf{x}_i\| \|\mathbf{x}_j\|} \right).
\end{equation}

The parameter $q$ is estimated according to $q = \max\left( \lfloor N_c / 20 \rfloor \right)$, where $N_c$ represents the
number of data points per cluster. $S_{1,2}$-LRR and $S_{2,3}$-LRR SC algorithms \cite{59} are low-rank
regularized SC methods, where low-rank constraints are implemented in terms of Schatten $S_{1,2}$
and $S_{2,3}$ norms. 

In relation to the proposed LIHFSS-SVDSC method, these linear SC algorithms
are inferior due to their shallow architecture, but they are superior due to dataset-dependent
tuning of hyperparameters. Thus, our reasoning is that this is a fair comparison. To achieve the best
performance, we tuned hyperparameters on ten randomly selected subsets using average
accuracy as the criterion. The number of samples per group for MNIST, USPS, Extended YaleB,
ORL, COIL20, and COIL100 datasets were 50, 50, 43, 7, 50, and 50, respectively. Clustering
performance was estimated as a mean value $\pm$ standard deviation of the corresponding metric
obtained by applying the tuned SC algorithms of 100 randomly generated partitions per group
as specified above. For these linear SC algorithms, we report in Tables 3 to 8 performance
without and with IPD post-processing with subspace dimensions known a priori, see section 2.1.

\subsection{Results}

The results of the comparative performance analysis across six datasets are presented in Tables
3 to 8. For the algorithms, which serve as oracles, the values of hyperparameters tuned on a held-out dataset are provided in parentheses. "RE" denotes the use of reconstruction
error loss during pre-training, while "DP" indicates distance-preserving loss was used. Results
incorporating IPD post-processing are included only when they led to performance improvements. 
\begin{table}[H]
    \centering
    \caption{\textbf{Clustering performance metrics on MNIST dataset.}}
    \renewcommand{\arraystretch}{1.2} 
    \setlength{\tabcolsep}{6pt} 
    \begin{adjustbox}{max width=\textwidth} 
    \begin{tabular}{l c c c}
        \toprule
        \textbf{Algorithm} & \textbf{ACC [\%]} & \textbf{NMI [\%]} & \textbf{F$_1$ score [\%]} \\
        \midrule
        LIHFSS-SVDSC RE & 57.50 & 55.99 & 49.82 \\
        LIHFSS-SVDSC RE+IPD & 59.40 & 55.38 & 49.41 \\
        ORACLE RE (0.1, 10)  & 63.15 & 68.01 & 56.70 \\
        LIHFSS-SVDSC DP & 53.98 & 56.16 & 49.24 \\
        LIHFSS-SVDSC DP+IPD & 60.27 & 66.01 & 55.56 \\
        ORACLE DP (0.1, 10)  & 62.50 & 59.00 & 56.72 \\
        \midrule
        SSC & 60.25$\pm$4.89 & 62.10$\pm$2.95 & 51.31$\pm$3.86 \\
        LRR SC & 54.46$\pm$5.40 & 60.71$\pm$3.27 & 45.78$\pm$4.35 \\
        LRR SC + IPD & 60.63$\pm$5.31 & 63.92$\pm$3.65 & 52.42$\pm$4.68 \\
        GMC LRSSC & 62.39$\pm$4.22 & 62.96$\pm$2.87 & 52.76$\pm$3.30 \\
        S0L0 LRSSC & 64.25$\pm$4.37 & 64.12$\pm$2.97 & 54.58$\pm$3.73 \\
        NSN & 68.82$\pm$5.02 & 66.73$\pm$3.76 & 58.52$\pm$4.68 \\
        RTSC & 60.49$\pm$3.70 & 61.81$\pm$2.96 & 51.48$\pm$3.68 \\
        RTSC+IPD & 61.94$\pm$4.69 & 64.95$\pm$3.17 & 54.12$\pm$4.19 \\
        LRR S$_{1/2}$ & 55.97$\pm$4.07 & 53.47$\pm$3.03 & 44.93$\pm$4.30 \\
        LRR S$_{1/2}$ + IPD & 58.40$\pm$5.03 & 59.84$\pm$3.41 & 49.17$\pm$4.31 \\
        LRR S$_{2/3}$ & 54.98$\pm$3.75 & 51.54$\pm$3.21 & 42.87$\pm$3.42 \\
        LRR S$_{2/3}$ + IPD & 56.40$\pm$3.63 & 54.75$\pm$3.08 & 45.60$\pm$3.46 \\
        \bottomrule
    \end{tabular}
    \end{adjustbox}
\end{table}
\begin{table}[H]
    \centering
    \caption{\textbf{Clustering performance metrics on USPS dataset.}}
    \renewcommand{\arraystretch}{1.2} 
    \setlength{\tabcolsep}{6pt} 
    \begin{adjustbox}{max width=\textwidth} 
    \begin{tabular}{l c c c}
        \toprule
        \textbf{Algorithm} & \textbf{ACC [\%]} & \textbf{NMI [\%]} & \textbf{F$_1$ score [\%]} \\
        \midrule
        LIHFSS-SVDSC RE & 78.72 & 80.14 & 78.45 \\
        LIHFSS-SVDSC RE+IPD & 80.82 & 82.82 & 79.16 \\
        ORACLE RE (100, 0.1)  & 81.72 & 83.02 & 79.34 \\
        LIHFSS-SVDSC DP & 75.82 & 79.85 & 59.93 \\
        ORACLE DP (100, 0.2)  & 81.93 & 83.28 & 79.17 \\
        \midrule
        SSC & 75.11$\pm$5.72 & 73.74$\pm$3.07 & 66.64$\pm$4.53 \\
        SSC+IPD & 75.03$\pm$5.77 & 74.76$\pm$3.57 & 67.80$\pm$5.17 \\
        LRR SC & 69.24$\pm$3.66 & 68.90$\pm$2.78 & 55.84$\pm$4.54 \\
        LRR SC + IPD & 80.77$\pm$5.67 & 80.76$\pm$2.41 & 74.18$\pm$4.34 \\
        GMC LRSSC & 79.47$\pm$4.76 & 76.91$\pm$2.46 & 71.25$\pm$3.56 \\
        GMC LRSSC + IPD & 78.44$\pm$5.18 & 78.37$\pm$2.65 & 71.86$\pm$3.98 \\
        S0L0 LRSSC & 82.75$\pm$6.14 & 80.57$\pm$3.09 & 75.47$\pm$5.03 \\
        NSN & 74.67$\pm$5.40 & 70.24$\pm$3.50 & 63.04$\pm$4.83 \\
        RTSC & 72.11$\pm$5.97 & 69.54$\pm$3.63 & 62.34$\pm$4.82 \\
        RTSC+IPD & 72.35$\pm$5.72 & 70.13$\pm$3.34 & 62.97$\pm$4.68 \\
        LRR S$_{1/2}$ & 77.24$\pm$3.49 & 69.18$\pm$2.54 & 63.63$\pm$3.29 \\
        LRR S$_{1/2}$ + IPD & 78.04$\pm$6.00 & 73.60$\pm$3.28 & 67.79$\pm$5.03 \\
        LRR S$_{2/3}$ & 75.78$\pm$3.92 & 68.67$\pm$2.53 & 62.55$\pm$3.22 \\
        LRR S$_{2/3}$ + IPD & 77.01$\pm$5.61 & 73.64$\pm$3.32 & 67.50$\pm$4.57 \\
        \bottomrule
    \end{tabular}
    \end{adjustbox}
\end{table}

\begin{table}[H]
    \centering
    \caption{\textbf{Clustering performance metrics on Extended YaleB dataset.}}
    \renewcommand{\arraystretch}{1.2} 
    \setlength{\tabcolsep}{6pt} 
    \begin{adjustbox}{max width=\textwidth} 
    \begin{tabular}{l c c c}
        \toprule
        \textbf{Algorithm} & \textbf{ACC [\%]} & \textbf{NMI [\%]} & \textbf{F$_1$ score [\%]} \\
        \midrule
        LIHFSS-SVDSC RE & 75.53 & 79.70 & 59.83 \\
        LIHFSS-SVDSC RE+IPD & 86.71 & 89.69 & 79.83 \\
        ORACLE RE (100, 0.2)  & 88.00 & 90.25 & 80.94 \\
        LIHFSS-SVDSC DP & 75.70 & 79.85 & 59.93 \\
        LIHFSS-SVDSC DP+IPD & 91.61 & 93.50 & 85.35 \\
        ORACLE DP (10, 0.01)  & 88.28 & 90.88 & 80.74 \\
        \midrule
        SSC & 75.65$\pm$1.84 & 80.43$\pm$1.65 & 40.77$\pm$4.39 \\
        SSC+IPD & 80.78$\pm$2.21 & 85.32$\pm$1.70 & 58.49$\pm$4.70 \\
        LRR SC & 73.38$\pm$2.59 & 83.10$\pm$1.34 & 48.22$\pm$5.67 \\
        LRR SC + IPD & 81.45$\pm$2.87 & 88.17$\pm$1.00 & 62.59$\pm$5.00 \\
        GMC LRSSC & 88.69$\pm$1.73 & 91.10$\pm$1.11 & 78.95$\pm$2.91 \\
        S0L0 LRSSC & 87.98$\pm$1.78 & 90.88$\pm$0.85 & 78.38$\pm$2.53 \\
        NSN & 72.98$\pm$2.53 & 75.35$\pm$1.46 & 57.72$\pm$2.51 \\
        RTSC & 40.50$\pm$1.62 & 52.26$\pm$1.17 & 20.53$\pm$1.38 \\
        LRR S$_{1/2}$ & 68.44$\pm$1.73 & 74.80$\pm$1.12 & 53.83$\pm$1.63 \\
        LRR S$_{1/2}$ + IPD & 91.20$\pm$1.10 & 92.82$\pm$0.49 & 84.04$\pm$1.52 \\
        LRR S$_{2/3}$ & 68.54$\pm$1.98 & 74.90$\pm$1.19 & 53.84$\pm$1.79 \\
        LRR S$_{2/3}$ + IPD & 91.56$\pm$1.55 & 92.93$\pm$0.61 & 84.25$\pm$1.71 \\
        \bottomrule
    \end{tabular}
    \end{adjustbox}
\end{table}
\begin{table}[H]
    \centering
    \caption{\textbf{Clustering performance metrics on ORL dataset.}}
    \renewcommand{\arraystretch}{1.2} 
    \setlength{\tabcolsep}{6pt} 
    \begin{adjustbox}{max width=\textwidth} 
    \begin{tabular}{l c c c}
        \toprule
        \textbf{Algorithm} & \textbf{ACC [\%]} & \textbf{NMI [\%]} & \textbf{F$_1$ score [\%]} \\
        \midrule
        LIHFSS-SVDSC RE & 75.50 & 86.12 & 66.21 \\
        LIHFSS-SVDSC RE+IPD & 78.00 & 87.26 & 67.08 \\
        ORACLE RE (0.0001, 0.1)  & 78.25 & 88.30 & 67.50 \\
        LIHFSS-SVDSC DP & 75.25 & 86.43 & 65.10 \\
        LIHFSS-SVDSC DP+IPD & 78.25 & 89.88 & 71.78 \\
        ORACLE DP (0.001, 10) & 78.50 & 87.36 & 68.18 \\
        \midrule
        SSC & 72.23$\pm$2.93 & 86.35$\pm$1.32 & 60.61$\pm$3.48 \\
        SSC+IPD & 74.74$\pm$2.71 & 87.62$\pm$1.23 & 63.84$\pm$3.25 \\
        LRR SC & 66.88$\pm$2.55 & 82.70$\pm$1.26 & 47.91$\pm$4.17 \\
        GMC LRSSC & 77.97$\pm$2.10 & 88.45$\pm$1.29 & 67.81$\pm$3.27 \\
        S0L0 LRSSC & 63.81$\pm$2.82 & 80.35$\pm$1.40 & 49.58$\pm$3.73 \\
        NSN & 67.80$\pm$2.52 & 82.78$\pm$1.27 & 53.11$\pm$3.12 \\
        RTSC & 69.12$\pm$2.70 & 82.86$\pm$1.40 & 54.38$\pm$3.34 \\
        LRR S$_{1/2}$ & 67.36$\pm$2.78 & 82.47$\pm$1.41 & 53.51$\pm$3.15 \\
        LRR S$_{2/3}$ & 68.00$\pm$3.00 & 82.88$\pm$1.47 & 54.63$\pm$3.45 \\
        \bottomrule
    \end{tabular}
    \end{adjustbox}
\end{table}
\begin{table}[H]
    \centering
    \caption{\textbf{Clustering performance metrics on COIL20 dataset.}}
    \renewcommand{\arraystretch}{1.2} 
    \setlength{\tabcolsep}{6pt} 
    \begin{adjustbox}{max width=\textwidth} 
    \begin{tabular}{l c c c}
        \toprule
        \textbf{Algorithm} & \textbf{ACC [\%]} & \textbf{NMI [\%]} & \textbf{F$_1$ score [\%]} \\
        \midrule
        LIHFSS-SVDSC RE & 68.61 & 76.38 & 62.39 \\
        LIHFSS-SVDSC RE+IPD & 82.50 & 89.19 & 78.07 \\
        ORACLE RE (0.5, 1000)  & 70.00 & 78.25 & 66.34 \\
        LIHFSS-SVDSC DP & 65.21 & 74.89 & 58.36 \\
        LIHFSS-SVDSC DP+IPD & 82.99 & 89.37 & 79.57 \\
        ORACLE DP (0.1, 1000)  & 69.24 & 77.03 & 62.99 \\
        \midrule
        SSC & 70.55$\pm$3.40 & 82.44$\pm$1.69 & 63.66$\pm$3.29 \\
        SSC+IPD & 75.12$\pm$3.17 & 85.45$\pm$1.53 & 69.34$\pm$3.31 \\
        LRR SC & 61.30$\pm$4.42 & 75.57$\pm$2.01 & 52.58$\pm$4.66 \\
        LRR SC+IPD & 70.34$\pm$3.43 & 82.29$\pm$1.96 & 62.73$\pm$4.19 \\
        GMC LRSSC & 71.45$\pm$2.70 & 82.98$\pm$2.41 & 63.46$\pm$3.35 \\
        S0L0 LRSSC & 69.93$\pm$2.99 & 81.14$\pm$1.54 & 62.48$\pm$2.90 \\
        S0L0 LRSSC + IPD & 70.28$\pm$2.77 & 81.37$\pm$1.78 & 62.95$\pm$3.11 \\
        NSN & 74.02$\pm$3.24 & 83.50$\pm$1.60 & 67.69$\pm$3.08 \\
        NSN+IPD & 77.75$\pm$2.64 & 84.60$\pm$1.39 & 71.01$\pm$2.70 \\
        RTSC & 72.51$\pm$3.29 & 82.23$\pm$1.49 & 65.04$\pm$3.21 \\
        LRR S$_{1/2}$ & 64.85$\pm$2.79 & 75.57$\pm$1.43 & 56.83$\pm$2.44 \\
        LRR S$_{1/2}$ + IPD & 66.65$\pm$3.26 & 77.48$\pm$1.54 & 57.91$\pm$3.17 \\
        LRR S$_{2/3}$ & 64.98$\pm$3.00 & 74.51$\pm$1.38 & 55.39$\pm$2.94 \\
        LRR S$_{2/3}$ + IPD & 68.55$\pm$3.70 & 79.07$\pm$1.85 & 59.95$\pm$3.81 \\
        \bottomrule
    \end{tabular}
    \end{adjustbox}
\end{table}
\begin{table}[H]
    \centering
    \caption{\textbf{Clustering performance metrics on COIL100 dataset.}}
    \renewcommand{\arraystretch}{1.2} 
    \setlength{\tabcolsep}{6pt} 
    \begin{adjustbox}{max width=\textwidth} 
    \begin{tabular}{l c c c}
        \toprule
        \textbf{Algorithm} & \textbf{ACC [\%]} & \textbf{NMI [\%]} & \textbf{F$_1$ score [\%]} \\
        \midrule
        LIHFSS-SVDSC RE & 52.03 & 75.67 & 46.67 \\
        LIHFSS-SVDSC RE+IPD & 50.66 & 79.66 & 45.10 \\
        ORACLE RE (0.1, 100)  & 55.00 & 78.38 & 48.12 \\
        LIHFSS-SVDSC DP & 51.30 & 75.59 & 45.65 \\
        LIHFSS-SVDSC DP+IPD & 50.47 & 79.72 & 44.94 \\
        ORACLE DP (0.5, 1000)  & 54.38 & 78.42 & 48.12 \\
        \midrule
        SSC & 51.17$\pm$1.33 & 78.08$\pm$0.69 & 41.84$\pm$1.75 \\
        SSC+IPD & 54.61$\pm$1.24 & 80.93$\pm$0.68 & 47.45$\pm$1.71 \\
        LRR SC & 36.84$\pm$2.30 & 69.20$\pm$1.83 & 16.79$\pm$4.15 \\
        LRR SC+IPD & 56.73$\pm$1.47 & 82.67$\pm$0.48 & 46.62$\pm$2.42 \\
        GMC LRSSC & 47.95$\pm$1.28 & 74.26$\pm$0.56 & 38.75$\pm$1.33 \\
        S0L0 LRSSC & 50.47$\pm$1.19 & 75.52$\pm$0.40 & 44.09$\pm$1.00 \\
        NSN & 57.15$\pm$1.11 & 79.88$\pm$0.43 & 49.73$\pm$1.05 \\
        RTSC & 59.87$\pm$1.63 & 84.37$\pm$0.41 & 50.58$\pm$1.70 \\
        LRR S$_{1/2}$ & 48.62$\pm$1.29 & 74.90$\pm$0.56 & 42.23$\pm$1.28 \\
        LRR S$_{1/2}$ + IPD & 48.96$\pm$0.95 & 77.65$\pm$0.41 & 42.08$\pm$1.21 \\
        LRR S$_{2/3}$ & 50.00$\pm$1.22 & 74.92$\pm$0.44 & 43.89$\pm$0.97 \\
        LRR S$_{2/3}$ + IPD & 51.20$\pm$1.00 & 77.77$\pm$0.42 & 44.84$\pm$0.84 \\
        \bottomrule
    \end{tabular}
    \end{adjustbox}
\end{table}
On the MNIST dataset, the LIHFSS-SVDSC method with DP loss and IPD post-processing achieved up to 3\% lower accuracy and 2\% lower NMI than the oracle using
reconstruction error, but 7\% higher NMI compared to the oracle using distance-preserving loss.
The F1-score was up to 1\% lower. Compared to the best linear SC algorithm, the NSN
algorithm, LIHFSS-SVDSC showed 8\% lower accuracy, less than 1\% lower NMI, and less than
3\% lower F1-score. Nevertheless, its performance was either comparable to or significantly
better than that of other linear SC algorithms. 

On the USPS dataset, the LIHFSS-SVDSC
method with RE loss and IPD post-processing performed within 1\% of the oracle DSC
algorithms in terms of accuracy, NMI, and F1-score. It also achieved comparable performance
to the S0L0-LRSSC algorithm and outperformed other linear SC algorithms. 

For the Extended YaleB
dataset, the our approach using DP loss and IPD post-processing outperformed
oracle DSC algorithms and all linear SC algorithms, demonstrating superior clustering
performance. On the ORL dataset, LIHFSS-SVDSC with DP loss and IPD post-processing
matched the accuracy of the oracle DSC algorithms while achieving almost 2\% higher NMI
and nearly 4\% higher F1-score. Additionally, it outperformed all linear SC algorithms, in some
cases by a considerable margin. 

For the COIL20 dataset, LIHFSS-SVDSC with DP loss and
IPD post-processing outperformed both oracle DSC algorithms and all linear SC algorithms, in
some cases by a large margin. On the COIL100 dataset, LIHFSS-SVDSC showed up to 3\%
lower accuracy, comparable or slightly lower NMI, and up to 3\% lower F1-score compared to
the oracle DSC algorithms. Compared to the best-performing linear SC algorithm, RTSC, our method exhibited 8\% lower accuracy, 5\% lower NMI, and 5\% lower F1-score.
To summarize the results of the comparative performance analysis, we emphasize that
the experimental results indicate that the LIHFSS-SVDSC method consistently performs close
to the oracle DSC methods across all six datasets. Regarding linear SC algorithms, LIHFSS-SVDSC achieved performance comparable to the best linear SC methods and outperformed
most of them. This outcome is attributed to the inherent limitations of linear SC methods, whose
objective functions and hyperparameters lead to dataset-dependent clustering performance. In
contrast, the proposed LIHFSS-SVDSC method, which integrates sequential pre-training and
fine-tuning, provides a more consistent and robust clustering performance across datasets.

\section{Ablation Studies}

The performance of LIHFSS-SVDSC is influenced by several key design choices, such as the selection of loss functions, the multi-stage training process, and post-processing techniques. To validate our methodology, we conducted two ablation studies.  Their purpose is to justify the proposed
methodology that performs learning in multiple stages sequentially, i.e., the pre-training stage
based on either reconstruction loss $\mathcal{L}_{\text{RE}}$ (13) or distance-preserving loss $\mathcal{L}_{\text{DP}}$ (17), and the fine-tuning stages based on layer-wise self-expression loss $\mathcal{L}_{\text{SE}}$ (18), and clustering quality loss $\mathcal{L}_{Q}$
(19)/(14). We present evidence which justifies: (i) use of $\mathcal{L}_{\text{SE}}$ (18) over $\mathcal{L}_{\text{SE}}^{M/2} =\|\mathbf{X^{M/2}} - \mathbf{X^{M/2}C}\|_F^2$
that is based on the encoder output only, (ii) the additional use of $\mathcal{L}_{Q}$ in the fine-tuning stage, and
(iii) use of IPD-based post-processing. Corresponding results for MNIST, ORL, and COIL20
datasets are presented in Tables 9 and 10. Performance metrics are given
in percentages.

\begin{table}[h]
    \centering
    \caption{\textbf{Ablation study on COIL20, MNIST, and ORL datasets with the $\mathcal{L}_{RE}$ in the pre-training stage}}
    \renewcommand{\arraystretch}{1.1} 
    \setlength{\tabcolsep}{4pt} 
    \begin{adjustbox}{width=\textwidth}
    \begin{tabular}{c c c c c | c c c | c c c | c c c}
        \toprule
        & & & & & \multicolumn{3}{c|}{\textbf{COIL20}} & \multicolumn{3}{c|}{\textbf{MNIST}} & \multicolumn{3}{c}{\textbf{ORL}} \\
        $\mathcal{L}_{RE}$ & $\mathcal{L}_{SE}^{M/2}$ & $\mathcal{L}_{SE}$ & $\mathcal{L}_{Q}$ & IPD 
        & ACC & NMI & F1 & ACC & NMI & F1 & ACC & NMI & F1 \\
        \midrule
        $\checkmark$ & $\times$ & $\times$ & $\times$ & $\times$ & 13.54 & 14.59 & 8.53 & 17.99 & 6.45 & 13.55 & 18.75 & 43.16 & 5.48 \\
        $\checkmark$ & $\checkmark$ & $\times$ & $\times$ & $\times$ & 62.08 & 75.88 & 50.78 & 28.12 & 15.82 & 17.93 & 54.25 & 65.82 & 47.93 \\
        $\checkmark$ & $\checkmark$ & $\times$ & $\checkmark$ & $\times$ & 63.00 & 76.62 & 52.07 & 29.50 & 17.42 & 19.82 & 55.00 & 65.24 & 48.27 \\

         $\checkmark$ & $\times$ & $\checkmark$ & $\times$ & $\times$ & 65.90 & 74.91 & 59.07 & 57.50 & 55.99 & 49.82 & 75.00 & 86.00 & 63.27 \\
        $\checkmark$ & $\times$ & $\checkmark$ & $\checkmark$ & $\times$ & 68.61 & 76.38 & 62.39 & 56.15 & 56.29 & 49.72 & 75.50 & 86.12 & 66.21 \\

        $\checkmark$ & $\times$ & $\checkmark$ & $\checkmark$ & $\checkmark$  & 82.50 & 89.19 & 78.07&59.40 & 55.38 & 49.41 & 78.00 & 87.26 & 67.08\\
        
        \bottomrule
    \end{tabular}
    \end{adjustbox}
\end{table}
\begin{table}[h]
    \centering
    \caption{\textbf{Ablation study on COIL20, MNIST, and ORL datasets with the $\mathcal{L}_{DP}$ in the pre-training stage}}
    \renewcommand{\arraystretch}{1.1} 
    \setlength{\tabcolsep}{4pt} 
    \begin{adjustbox}{width=\textwidth}
    \begin{tabular}{c c c c c | c c c | c c c | c c c}
        \toprule
        & & & & & \multicolumn{3}{c|}{\textbf{COIL20}} & \multicolumn{3}{c|}{\textbf{MNIST}} & \multicolumn{3}{c}{\textbf{ORL}} \\
        $\mathcal{L}_{DP}$ & $\mathcal{L}_{SE}^{M/2}$ & $\mathcal{L}_{SE}$ & $\mathcal{L}_{Q}$ & IPD 
        & ACC & NMI & F1 & ACC & NMI & F1 & ACC & NMI & F1 \\
        \midrule
        $\checkmark$ & $\times$ & $\times$ & $\times$ & $\times$ & 16.66 & 12.94 & 8.57 &16.92&3.52&11.75&17.75 & 41.83 & 2.91 \\
        $\checkmark$ & $\checkmark$ & $\times$ & $\times$ & $\times$ &60.17&71.23&48.42&21.01&8.54&14.13&57.25 & 75.06 & 43.85\\
        $\checkmark$ & $\checkmark$ & $\times$ & $\checkmark$ & $\times$ & 58.06 &75.19 & 51.39 & 45.06&49.19&37.39&59.25&76.17&45.62\\

        $\checkmark$ & $\times$ & $\checkmark$ & $\times$ & $\times$ &60.56&72.46&53.35&55.69&52.85&46.49&70.75 & 83.63 & 61.66  \\
        $\checkmark$ & $\times$ & $\checkmark$ & $\checkmark$ & $\times$ & 65.21 & 74.89&58.36&53.98&56.16 & 49.24 & 75.25& 86.43& 65.10 \\
       
        $\checkmark$ & $\times$ & $\checkmark$ & $\checkmark$ & $\checkmark$ & 82.99 & 89.37 & 79.57& 60.27 & 66.01 & 55.56 & 78.25 & 89.88 & 71.78 \\
        \bottomrule
    \end{tabular}
    \end{adjustbox}
\end{table}

\section{Conclusion}

Due to their strong representation learning capabilities, neural networks emerged as a powerful tool for clustering data from nonlinear manifolds. The dominant approach involves
applying linear subspace clustering algorithms to the embedded data at the encoder's output,
where the learned representations are assumed to follow the union-of-subspaces model.
However, this approach overlooks valuable information from intermediate layers and does not
allow clustering performance to influence the representation learning process. Recent efforts to
address these limitations have resulted in self-supervised deep subspace clustering networks,
also known as self-guided or self-taught networks, which incorporate composite loss
functions as weighted combinations of multiple loss terms. However, these methods implicitly
assume access to a held-out dataset for hyperparameter tuning, a condition that is often
unrealistic in real-world applications. 

To overcome these challenges, we proposed a fully label
independent deep subspace clustering methodology that does not require hyperparameter
tuning. Our key innovation is a sequential learning framework, consisting of a pre-training stage
followed by fine-tuning with carefully selected loss functions that reflect the intrinsic properties
of the data. This approach eliminates the need for manual hyperparameter selection and ensures
applicability to real-world datasets without requiring a held-out validation set—a major
advantage of the proposed methodology. Despite its strengths, the sequential learning procedure
has limitations compared to simultaneous optimization of multiple weighted loss functions. Our
ablation studies indicate that although minimizing clustering quality loss after layer-wise self-expression loss leads to performance improvements, the gains are marginal. Consequently,
even though our methodology demonstrates competitive results in comparative performance
analysis, its sequential nature remains its primary limitation. In future work, we aim to develop
a fully label-independent deep subspace clustering framework that optimizes multiple weighted
loss functions simultaneously. We plan to explore various strategies for generating a target
distribution and subsequently select hyperparameters that best align with it. Alternatively, we
will investigate internal clustering quality metric-based approaches, focusing on intra-cluster compactness and inter-cluster separability, to further enhance clustering performance.

\section*{CRediT authorship contribution statement}

\textbf{Lovro Sindičić:} Writing – review \& editing, Writing – original draft, Visualization, Validation,
Resources, Methodology, Investigation. \textbf{Ivica Kopriva:} Writing – review \& editing,
Supervision, Project administration, Funding acquisition.

\section*{Declaration of Competing Interests}

We declare that this manuscript is original, has not been published before and is not currently
being considered for publication elsewhere. We declare that we do not have any known
competing financial interests or personal relationships that could have appeared to influence
the work reported in this paper.

\section*{Declaration of generative AI and AI-assisted technologies in the writing process}

During the preparation of this work the author used \textit{ChatGPT 4} in order to improve the
readability and language of the paper.

\section*{Data Availability}

Data used in reported experiments are available either at links cited in the paper or at:
\url{https://github.com/lovro-sinda/LIHFSS-SVDSC}.

\section*{Acknowledgments}
This work was supported by the Croatian Science Foundation under project number HRZZ-IP-2022-10-6403.

\bibliographystyle{elsarticle-num}

\end{document}